# Temporal-Spatial dependencies ENhanced deep learning model (TSEN) for household leverage series forecasting

Hu Yang[1,§], Yi Huang[1], Haijun Wang[2,*], and Yu Chen[3]

([1] School of Information, Central University of Finance and Economics, Beijing, 100081, China; [2] School of Economics, Beijing Wuzi University, Beijing, 102600, China; [3] School of Public Finance and Taxation, Central University of Finance and Economics, Beijing, 100081, China)

**Corresponding authors**: [*]Haijun Wang (wanghaijun2005@126.com), [§] Hu Yang (hu.yang@cufe.edu.cn)

**Abstract**

Analyzing both temporal and spatial patterns for an accurate forecasting model for financial time series forecasting is a challenge due to the complex nature of temporal-spatial dynamics: time series from different locations often have distinct patterns; and for the same time series, patterns may vary as time goes by. Inspired by the successful applications of deep learning, we propose a new model to resolve the issues of forecasting household leverage in China. Our solution consists of multiple RNN-based layers and an attention layer: each RNN-based layer automatically learns the temporal pattern of a specific series with multivariate exogenous series, and then the attention layer learns the spatial correlative weight and obtains the global representations simultaneously. The results show that the new approach can capture the temporal-spatial dynamics of household leverage well and get more accurate and solid predictive results. More, the simulation also studies show that clustering and choosing correlative series are necessary to obtain accurate forecasting results.

**Acknowledgment**

HY was supported by grants from the National Natural Science Foundation for Distinguished Young Scholars of China (71701223), the National Statistical Science Foundation of China (2018LZ08), the Central University of Finance and Economics Young Talents Training Support Project (QYP2014), and Fundamental Research Funds for the Central Universities (China): the Central University of Finance and Economics Scientific Research and Innovation Team Support Project. YC was supported by grants from the National Social Science Foundation (20BGL066) and the "Young Talents" Cultivation and Support Program of the Central University of Finance and Economics.

# Temporal-Spatial dependencies ENhanced deep learning model (TSEN) for household leverage series forecasting

**Abstract**

Analyzing both temporal and spatial patterns for an accurate forecasting model for financial time series forecasting is a challenge due to the complex nature of temporal-spatial dynamics: time series from different locations often have distinct patterns; and for the same time series, patterns may vary as time goes by. Inspired by the successful applications of deep learning, we propose a new model to resolve the issues of forecasting household leverage in China. Our solution consists of multiple RNN-based layers and an attention layer: each RNN-based layer automatically learns the temporal pattern of a specific series with multivariate exogenous series, and then the attention layer learns the spatial correlative weight and obtains the global representations simultaneously. The results show that the new approach can capture the temporal-spatial dynamics of household leverage well and get more accurate and solid predictive results. More, the simulation also studies show that clustering and choosing correlative series are necessary to obtain accurate forecasting results.

## 1 Introduction

Time series forecasting (TSF) is imperative to a wide range of financial forecasting problems that have a temporal pattern. For instance, with the help of forecasting tools, if the governors of a country can foresee that their nation might suffer from financial risk in the next couple of months, they will make a good fiscal policy that allocates sufficient resources to hedge against market risks and optimize investments in advance. Such financial risks may be caused by the rapid raising household debt, which always amplifies downturns, weakens recoveries, and serves as the fuse for an outbreak of financial crisis (Clarke, 2019; Mian, Sufi, & Verner, 2017), or the drastic fluctuating international exchange rate(Ca'Zorzi & Rubaszek, 2020). Due to the complex and continuous fluctuation of impacting factors, real-world time series tend to be extraordinarily non-stationary, which exhibit diverse dynamics. For example, the household debt (Verner & Gyngysi, 2020) of a certain region is largely affected not only by exogenous variables, but also by the location of the region. The location is representing the spatial pattern, where similar series could have similar trends, variations, and uncertainty. Another example is the international exchange rate (Ca'Zorzi & Rubaszek, 2020), which is influenced by both the domestic economy and economies of many associated countries. It also has diverse dynamical patterns: the temporal pattern within a specific series and the spatial correlation pattern among the target series and its associated series. In this work, we will study multiple multi-variate time series forecasting: multi-variate time series evolve with time; and, they are spatially correlated.

Many traditional statistical-based models and machine learning models have been developed for computers to model and learn the trend and seasonal variations of the series and also the correlation between observed values that are close in time. For instance, the autoregressive integrated moving average (ARIMA) (Saboia, 1977; Tsay, 2000), as a classical linear model in statistics, is an expert in modeling and learning the linear and stationary time dependencies with a noise component (De Gooijer & Hyndman, 2006), the multivariate autoregressive time series models (MAR) (Fountis & Dickey, 1989) that can learn time series patterns accompanied by explanatory variables. Moreover, several statistical methods have been developed to extract the nonlinear signals from the series, such as the bilinear model (Poskitt & Tremayne, 1986), the threshold autoregressive model (Stark, 1992), and the autoregressive conditional heteroscedastic (ARCH) model (Engle, 1982). However, these models have a rigorous requirement of the stationarity of a time series, which encounters severe restrictions in practical use if most of the impacting factors are unavailable.

Since the time series prediction is closely related to regression analysis in machine learning, traditional machine learning models (MLs), such as decision tree (DT) (Galicia, Talavera-Llames, Troncoso, Koprinska, & Martínez-Álvarez, 2019; Lee & Oh, 1996), support vector machine (SVM), and k nearest neighbor (kNN), can be used for time series forecasting (Galicia, et al., 2019). Inspired by the notable achievements of deep learning (DL) in natural language processing (Devlin, Chang, Lee, & Toutanova, 2018), image classification (Krizhevsky, Sutskever, & Hinton, 2012), and reinforcement learning (Silver, et al., 2016), several artificial neural network (ANN) algorithms have drawn people's attention and become strong contenders along with statistical methods in the forecasting community with their better prediction accuracies (Zhang, Patuwo, & Hu, 1998). Significantly, different from MLs that require hand-crafted features, DLs have a great potential to learn complex non-linear temporal feature interactions among multiple series. Because DLs automatically learn complex data representations of an MTS, they alleviate the need for manual feature engineering and model design (Bengio, Courville, & Vincent, 2013; Lim & Zohren, 2021). Moreover, DLs can learn the linear and nonlinear patterns of data better.

Initially, most DLs are developed to model and learn the temporal dependency of time series. For instance, the simplest DL, the recurrent neural network (RNN), can store a lot of information about the past and it allows updates of its hidden state dynamically (Rumelhart et al. 1986; Werbos 1990; Elman 1990). To address the weakness of RNNs in managing long-term dependencies, the long-short term memory (LSTM) (Hochreiter & Schmidhuber, 1997), a variant of RNN capable of learning long-term dependence, has also been employed for series forecasting (Gers, Schmidhuber, & Cummins, 2000). LSTM comprises a separate autoencoder and forecasting sub-models. LSTM has an RNN architecture but it is different from RNN, whereas it can solve the problem of vanishing gradient. The Gate Recurrent Unit (GRU) (Dey & Salem, 2017) is also an important variant of RNN, where its basic idea of learning long-term dependence is consistent with LSTM; however, it only uses a reset gate and an update gate. The long- and short-term time-series network (LSTNet) (Lai, Chang, Yang, & Liu, 2018) is designed specifically for MTS forecasting with up to hundreds of time series. LSTNet uses CNNs to capture short-term patterns and LSTM (Hochreiter & Schmidhuber, 1997) or GRU (Dey & Salem, 2017) for memorizing relatively long-term patterns. Besides, the attention mechanism (Bahdanau, Cho, & Bengio, 2014; Luong, Pham, & Manning, 2015), originally utilized in encoder-decoder networks (Krizhevsky, et al., 2012), somewhat solves the problem of integrating correlative unites, and thus increases the effectiveness of RNNs (Lai, et al., 2018). The temporal pattern attention reviews the information at each stage and selects relevant information to help to generate the outputs (Shih, Sun, & Lee, 2019). Recent studies demonstrate how both the automatic feature learning capabilities of LSTMs and their ability to handle input sequences can be harnessed in an end-to-end model that can be used to drive demand forecasting (Hu & Zheng, 2020).

Besides learning the dynamics of temporal dependence, time series that exhibit spatial dependencies are also important information of time series. The spatio-temporal (ST) properties are commonly observed in various fields, such as transportation (Shao, Salim, Gu, Dinh, & Chan, 2017), social science (Kupilik & Witmer, 2018), and criminology (Rumi, Luong, & Salim, 2019). Some researchers have made efforts to utilize spatial correlation of multiple target time series to realize accurate forecasting. In statistics, the fully Spatio-temporal MAR (ST-MAR) model is developed within the framework of functional data analysis to utilize both the linear temporal patterns of the series itself and the linear spatial patterns of its neighbors (Valdes-Sosa, 2004). Although ST-MAR is doing well in the inclusion of spatial information, ST-MAR has the same problems while analyzing nonlinear and non-stationary time series similar to MAR. Similarly, spatio-temporal modeling has seldom been taken into account in the DLs, and DLs models consist of two components: one is for capturing the spatio-temporal dynamical pattern of the series; and the other one is for decoding these latent states and translating them into actual series observations. Based on the design, models can capture the dynamics and correlations in multiple series at the spatial and temporal levels (Ziat, Delasalles, Denoyer, & Gallinari, 2017). For instance, PV energy production prediction (Ceci, Corizzo, Fumarola, Malerba, & Rashkovska, 2016), traffic time series forecasting (Cirstea, Yang, Guo, Kieu, & Pan, 2022), covid-19 forecasting (Kapoor, et al., 2020), and brain-computer interface (BCI) (Topic & Russo, 2021), all of which are both spatial and temporal dependencies. Therefore, they demonstrate good performance on forecasting tasks.

Although DLs are state-of-art techniques and good for modeling and learning the nonlinear and non-stationary time series with spatial patterns, implementation of DLs in forecasting financial time series projects would not provide significant improvement in forecasting. On the one hand, while DLs were successful in some instances, where the series being extrapolated are often numerous and long, in typical time series forecasting, where data is insufficient and the regressor is unavailable, the performance of DLs algorithms tends to be under expectations (Makridakis, Spiliotis, & Assimakopoulos, 2018). For instance, some finance time series, like household debt, are short in time with limited observations. On the other hand, both the spatial proximity and the long-term temporal correlations of the data are usually complex and hard to be captured. Moreover, previous spatial-temporal methods assume neighboring individuals interfere with each other, so they learn the representation of spatial correlation based on the given graph structure. For instance, the neighbor pixels usually have similar RGB values in image and video (Topic & Russo, 2021), and adjacent nodes in the road may cause congestion one after the other (Cirstea, Yang, Guo, Kieu, & Pan, 2022). However, in financial time series, the structural relationship between any two individual time series is uncertain. Meanwhile, a series spatially depends on which time series is also unknown. These factors would impede the way of utilizing spatial patterns to enhance the performance of the forecasting models.

With the recent advancements in DLs techniques, we are now capable of handling complex dynamics as a single unit, even without any additional impact factors. In this paper, we study forecasting models in both a short series and a long series in finance – focusing on the key example of the household debt and international exchange rate – in a data-rich environment, where our data includes not only conventional multi-variate series but also multiple target time series. We find that our forecasts are either superior to or as good as those benchmark DLs. This is the case when (a) we compare our approach with the CNN, LSTM, and GRU in terms of forecasting the series of household debt and the series of international exchange rates or (b) we compare our approach with other models in the artificial data. The former is a comparison of different methods, whereas the latter reveals under which conditions the model could perform well. In addition, we also conduct statistical testing to evaluate the difference between the new method and previous DLs.

We make several novel contributions to the new model to achieve our goal. (1), a new method, the **T**emporal-**S**patial dependencies **EN**hanced deep learning model (**TSEN**), is proposed to forecast the short and long financial time series. The method consists of two components: one captures new representations of spatio-temporal dynamics of the series, and another one decodes these representations into target series observations. It is finally used to forecast the household leverage in multiple regions and the international exchange rate of multiple countries simultaneously. (2) The accuracy and robustness of the proposed approach are validated through applications of forecasting multiple MTS. (3) The model is also validated by simulated datasets to explain under which conditions it could outperform previous DLs.

The rest of the paper is organized as follows. Section 2 presents the related studies on time series analysis. Section 3 presents the issue and notations of our studies. In section 4, we describe the framework of the Temporal-Spatial dependencies ENhanced deep learning model (TSEN). Section 5 describes two financial time series and the way of generating artificial data. Section 6 elaborates on the experimental results of forecasting time series in the previous section. Finally, we provide the concluding remarks in Section 7.

## 2. Preliminary

The goal of time series forecasting is to predict its value at $t+h$ based on available observations from a time series at time $t$. Suppose if there is only one single time-dependent variable is available, the problem can be studied using univariate time series (UTS) analysis methods, formulated as

$$\hat{y}_{t+h} = f(y_t, y_{t-1}, \ldots, y_{t-k}; \theta) \tag{1}$$

where $y_t, y_{t-1}, \ldots, y_{t-k}$ refers to time series data points, $\theta$ are the parameters such as autoregression coefficients, $\hat{y}_{t+h}$ is the forecasting values at $t+h$, $k$ is the number of inputs, and $h = 1,2,\ldots$ is any positive integer. For instance, ARIMA and its variants can model and learn stationary UTS well. With some exogenous time series data, the problem of financial time series forecasting turns into multivariate time series (MTS) analysis, which can be formulated as

$$\hat{y}_{t+h} = f(Y_t, X_t; \theta) \tag{2}$$

where $Y_t = (y_t, y_{t-1}, \ldots, y_{t-k})$ refers to the target time and $X_t = (X_{1t}, X_{2t}, \ldots, X_{mt})$ is the exogenous MTS whose item is $X_{it} = (x_{i,t}, x_{i,t-1}, \ldots, x_{i,t-k})$ for $i = 1,2,\ldots,m$. Financial time series forecasting is a type of MTS analysis, which can be implemented by both traditional methods and state-of-art deep learning methods, such as RNN (Rumelhart et al. 1986; Werbos 1990; Elman 1990), LSTM (Hochreiter & Schmidhuber, 1997), GRU (Dey & Salem, 2017), and so on.

However, in practical circumstances, such as household debt(Verner & Gyngysi, 2020), international exchange rate(Ca'Zorzi & Rubaszek, 2020), cryptocurrency(Chen, Xu, Jia, & Gao, 2021), retail sales(Rafiei & Adeli, 2016), and energy consumption(Deb, Zhang, Yang, Lee, & Shah, 2017), datasets are collected as spatially indexed MTS and are often spatially correlated because of their similar location, or economic structures, or development levels. This fact indicates that the variances of a target series may be influenced by others. The inclusion of spatial dependencies in the forecasting model may enhance the performance of the model. We use $\{Y_{j,t}, X_{j,t}\}$ to denote a MTS for the $j$th region or nation, where $j = 1,2,\ldots,J$ and $J$ is the number of regions or nations we have observed. Thus, the problem of modeling multiple MTSs $\{Y_{j,t}, X_{j,t}\}_{j=1}^{J}$ to forecast multiple target time series is formulated as

$$\hat{y}_{1,t+h}, \hat{y}_{2,t+h}, \ldots, \hat{y}_{J,t+h} = f(\{Y_{j,t}, X_{j,t}; \theta_j\}_{j=1}^{J}; \gamma) \tag{3}$$

where $\theta_j$ describes the relationship between $Y_{j,t}$ and $X_{j,t}$ for an MTS, and $\gamma$ represents the relationship between an MTS and another. Formula (3) is the forecasting model with multiple responses or outputs. The common limitation of RNN, LSTM, GRU, and their extensions is that they inadequately deal with multiple MTS with spatial correlation to some extent. To overcome the weakness, some recent spatial-temporal models rely on the graph structure, which describes the spatial dependencies of series, but still has the problem that the graph structure is sometimes unknown, which hinders the use of spatial dependencies in the process of designing forecasting models. For the above-mentioned reasons, we propose a new accurate and stable forecasting model based on the potential application of TSF in finance and take China's household leverage and international exchange rates as examples. Our study will help policymakers to reasonably evaluate the changes in financial time series and evaluate the risk ahead of time, and then provide support for reasonably controlling the financial risk and policy intervention. The goal of this study is to develop an end-to-end forecast model for multiple multi-step MTS forecasting tasks that handle multiple MTS inputs. Finally, we want to answer the following three questions in this paper.

Q1: How to model and learn the temporal-spatial patterns of multiple MTSs?

Q2: (1) How to choose or screen the correlated multiple series MTSs and implement the forecasting model? (2) Can the spatial dependencies make better performance of prediction?

Q3: If the answer to Q2 is "yes," is it better to include many more series for forecasting than just a few series?

## 3. Related studies

### 3.1 LSTM Layer

The long short-term memory (LSTM) can automatically learn the representation of MTS and then harness the embeddings in an end-to-end model that can be used to drive demand forecasting (Hu and Zheng, 2020). Fig. 1 shows the structures of two canonic RNN-based methods: LSTM and GRU.

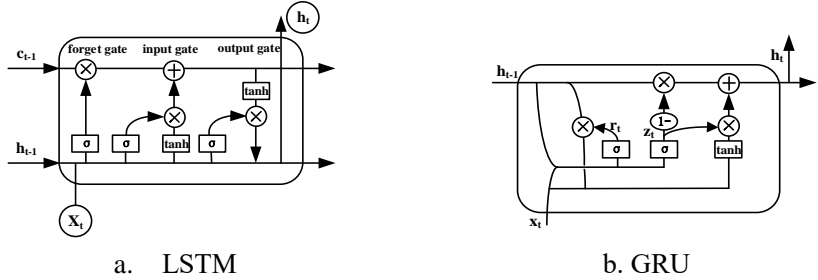

a. LSTM            b. GRU

Fig. 1. Structures of LSTM and GRU

LSTM (Hochreiter & Schmidhuber, 1997) is capable of learning long-term dependencies while the RNN (Rumelhart et al. 1986; Werbos 1990; Elman 1990) captures the temporal dependencies from the historical UTS or MTS. LSTM consists of four components as shown in Fig. 1 (subgraph a): the forget gate, the input gate, the output gate, and the unit status, which are formulated as

$$f_t = \sigma(W_f \cdot [h_{t-1}, X_t] + b_f) \tag{4}$$
$$j_t = \sigma(W_j \cdot [h_{t-1}, X_t] + b_j) \tag{5a}$$
$$\tilde{c}_t = tanh(W_c \cdot [h_{t-1}, X_t] + b_c) \tag{5b}$$
$$c_t = f_t * c_{t-1} + j_t * \tilde{c}_t \tag{6}$$

$$o_t = \sigma(W_o \cdot [h_{t-1}, X_t] + b_o) \qquad (7a)$$
$$h_t = o_t * \tanh(c_t) \qquad (7b)$$

where $W_f$, $W_j$, $W_c$, $W_o$, $b_f$, $b_j$, $b_c$ and $b_o$ are the trainable parameters, $X_t$ is the input MTS, and $h_t$ is the representation of $X_t$ learned by the LSTM at time t.

The first component is the so-called "*forget gate layer*" in Formula (4), which determines which information needs to be thrown away from the cell state; the decision is made by a sigmoid operation. It takes $h_{t-1}$ and $X_t$ as input, and outputs a number between 0 and 1 for each number in the cell state $c_{t-1}$; here, 1 represents "completely keep this," while 0 represents "completely get rid of this." The second component is the decoder, which produces the target outputs organized by time steps from a context vector generated by the encoder. The next component decides which new information needs to be stored in the cell state. It contains two parts: one is a sigmoid layer in Formula (5a), also named the "input gate layer," which decides which values will be updated; the other is the $\tanh(\cdot)$ operation, which creates a vector of new candidate values $\tilde{c}_t$ that can be added to the state in Formula (5b). The third component changes the old cell state $c_{t-1}$ into a new cell state $c_t$. Since the previous component has already decided what to do, this component just multiplies the old state $c_{t-1}$ by $j_t$ to discard the information that has been decided to be forgotten earlier. Then, we add the new candidate values $j_t * \tilde{c}_t$, which are scaled by how much the state value needs to be updated. The operation is defined in Formula (6). The last component is the output layer. The output will be based on the cell state but will be a filtered version. First, a sigmoid layer is used to decide which parts of the cell state need to output in Formula (7a). Then, the cell state is put through a $\tanh(\cdot)$ layer (to push the values to be between −1 and 1) and multiply it by the output of the sigmoid gate in Formula (7b). In summary, the LSTM is used to learn a new representation of MTS for forecasting can be simplified and re-arranged as

$$H_t = \text{LSTM}(Y_t, X_t, w) \qquad (8)$$

where w is trainable parameters of LSTM, and $H_t$ is the new representation of $Y_t, X_t$.

*3.2 GRU Layer*

GRU (Dey & Salem, 2017) is a simple version of LSTM, as it uses the same gate to carry out forget and select memory simultaneously. It has fewer parameters and provides competitive performance over LSTMs. Compared to LSTM model, GRU decreases the number of gates from three to two, where the two gates are called updated gate $z_t$ and a reset gate $r_t$. The GRU model is formulated as

$$r_t = \sigma(W_r \cdot [h_{t-1}, X_t] + b_r) \qquad (9)$$
$$z_t = \sigma(W_z \cdot [h_{t-1}, X_t] + b_z) \qquad (10)$$
$$\tilde{h}_t = tanh(W \cdot [r_t * h_{t-1}, X_t] + b_c) \qquad (11)$$
$$h_t = (1 - z_t) * h_{t-1} + z_t * \tilde{h}_t \qquad (12)$$

where $W_z$, $W_r$, W, $b_z$, $b_r$, and $b_c$ are the trainable parameters, $X_t$ is the input MTS, and $h_t$ is the representation of $X_t$ learned by the GRU at time t, $\sigma(\cdot)$ is the sigmoid function. The first component acts as the "*reset gate*" that determines which parts of the previous hidden state need to be considered, or ignored at the current operation in (9). The second component is the so-called "*update gate*" that determines which parts of the previous memory need to be updated and changed to the new candidate memory in (10). The third component computes the candidate state at the current step using the previous hidden state, the output of the *reset gate* $r_t$, and the input $X_t$ in (11). Operation $r_t * h_{t-1}$ determines which hidden states will be preserved for the candidate state. The final one is used to obtain the representation of both $h_{t-1}$ and $\tilde{h}_t$: if $z_t$ is closer to 1, then more data will be memorized; while if it is closer to 0, then more data will be forgotten in (12). Similarly, the GRU is also used to learn a new representation of MTS for forecasting tasks, which is

$$H_t = \text{GRU}(Y_t, X_t, \text{w}) \tag{13}$$

*3.3 Attention Layer*

In neural networks, attention (Vaswani, et al., 2017) is a technique that mimics human cognitive attention. It is inspired by humans' biological mechanism, where a person tends to concentrate on the important things while the brain is processing large amounts of information, among which the brain only picks the important things as needed. Thus, the effects of some parts of the input are enhanced while other parts are weakened. This phenomenon leads to the idea that the neural network should focus on the small, but important, parts of the input. Learning which part of the input is more important than others depends on the context. Let the input be $\{h_1, h_2, \ldots, h_m\}$, which may be the original data or the output of the neurons in the networks. Considering the fact that various inputs may play different roles in the process of forecasting, the attention layer is used to learn the attention presentation, which is formulated as

$$\alpha_{ts} = \frac{\exp(\text{score}(h_t, \bar{h}_s))}{\sum_{s'=1}^{m} \exp(\text{score}(h_t, \bar{h}_{s'}))} \tag{14}$$

$$c_t = \sum_s \alpha_{ts} \bar{h}_s \tag{15}$$

$$\boldsymbol{a}_t = f(c_t, h_t) = \tanh(\boldsymbol{W}_a[h_t, c_t]) \tag{16}$$

where $h_t$ and $\bar{h}_s$ are the input, $\alpha_{ts}$ is the attention weight, $c_t$ is the linear combination of $\bar{h}_s$, $\boldsymbol{W}_a$ are the learnable parameters, $\boldsymbol{a}_t$ is the output of the attention layer, score(·) is a function used to compute the similarity of $h_t$ and $\bar{h}_s$. In our study, $h_t$ and $\bar{h}_s$ are the same. And, the attention layer is used to learn the new representation for forecasting the *j*th MTS, which is the combination of representations of all MTS related to the target series. The simplified expression is

$$H_{j,a} = \text{Attention}\left(\{H_{j,t}\}_{j=1}^{J}; \gamma\right) \tag{17}$$

where $\gamma$ is the attention weight.

## 4. The new method

In this section, we answer Q1 and Q2 by demonstrating the implementation of a new framework for leaning and modeling multiple MTSs step by step. Multiple MTSs usually contain both temporal and spatial patterns that are important to forecasting. The temporal patterns represent the trend of a specific MTS while the spatial patterns reveal the correlation between one MTS and another. Inspired by the success of the deep learning model, we propose the **T**emporal-**S**patial dependencies **EN**hanced deep learning model (**TSEN**) to forecast the financial time series by utilizing both temporally and spatially correlated information. The framework of the new approach shown in Fig. 2 consists of two critical steps as other deep learning methods. Step one is to capture the temporal-spatial dynamics of the process and obtain latent states or representations. It contains multiple RNN-based layers and multiple attention layers, in sense of that both global (applicable to extract spatial dependence of all series) and local (applied to learn the representation of each series individually) parameters are utilized in order to enable cross-learning while also emphasizing the particularities of the time series being extrapolated. Step two is to transform the latent representation into the output. For instance, in the first step, the network needs to learn both global and local representations of $\{Y_{j,t}, \mathbf{X}_{j,t}\}_{j=1}^{J}$, obtain latent states $\{Z_j\}_{j=1}^{J}$, and take $\{Z_j\}_{j=1}^{J}$ as inputs to implement the forecasting model $\hat{y}_{1,t+h}, \hat{y}_{2,t+h}, \ldots, \hat{y}_{J,t+h} = f(\{Z_j\}_{j=1}^{J})$. More details of the TSEN model can be found in Fig. 1.

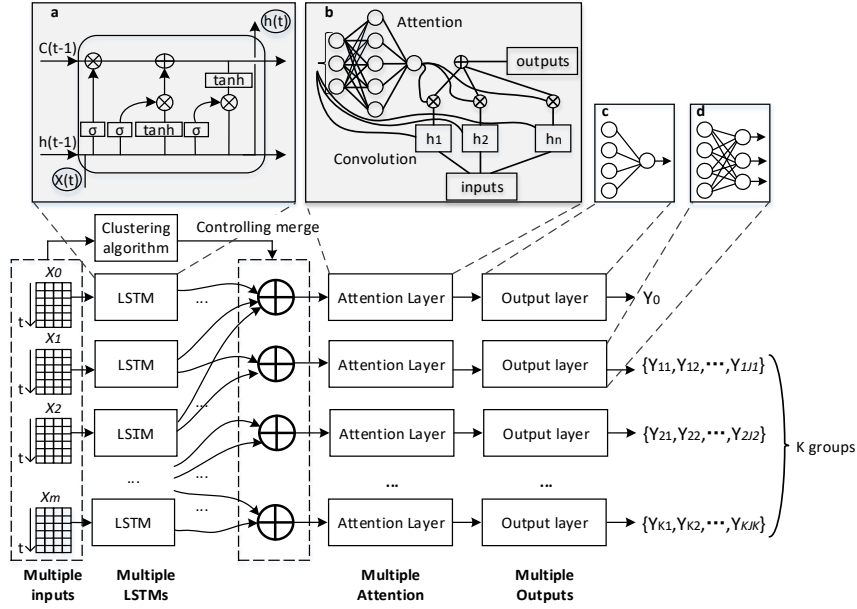

Fig.2. Framework of **T**emporal-**S**patial dependencies **EN**hanced deep learning model (**TSEN**), where **a** is the LSTM layer, **b** is the attention layer, and **c** and **d** are the output layer or the feedforward neural network.

*4.1 Clustering and screening*

Although including multiple MTSs may enhance the performance of forecasting, redundant MTSs may cause overfitting if the sample size is limited. For this reason, we only add the MTSs that are highly correlated to the target series which is needed for prediction in the model. To screen these MTS, which can be used to predict the target series, we perform the flowing steps. Firstly, we calculate the similarity of any two UTS of the target series by using the Euclid distance to estimate their correlations. For example, given any two UTS of household leverage $Y_u$ and $Y_v$, their similarity is defined as $d_{uv} = \|Y_u - Y_v\|_2^2$. We prefer the Euclid distance to the correlation coefficient because of the significant difference in the scale of financial time series in different regions. Secondly, after computing the similarities, the hierarchical clustering algorithm is employed to divide $J$ MTSs into different groups. It means that the sequence of multiple MTSs $\{Y_{1,t}, Y_{2,t}, \ldots, Y_{J,t}\}$ can be grouped into $K$ sections. Take the $g$th group as an example: it contains $g_J$ MTSs, denoted as $\{Y_{g_j,t}, \mathbf{X}_{g_j,t}\}_{g_j=1}^{g_J}$, $g_j$ is the index; for any two clusters $g$th and $q$th ($g \neq q$), $\{Y_{g_j,t}, \mathbf{X}_{g_j,t}\}_{g_j=1}^{g_J} \cap \{Y_{q_j,t}, \mathbf{X}_{q_j,t}\}_{q_j=1}^{q_J} = \emptyset, g, q = 1, \ldots, K$, and $\sum_{g=1}^{K} g_J = J$.

*4.2 Learning representation from temporal patterns*

To model and learn the temporal pattern of multiple MTS, we take a set of MTSs $\{Y_{g_j,t}, \boldsymbol{X}_{g_j,t}\}_{g_j=1}^{g_J}$ as input series, and then separately conduct parallel computing using multiple LSTMs to extract useful information from the set of MTSs sequentially. Take LSTM(·) as an example: $\{h_{1,t}, \dots, h_{J,t}\}$ is the latent variables; here, a multiple LSTM is defined as follows.

$$\begin{aligned} h_{1,t} &= \text{LSTM}(Y_{g_1,t}, \boldsymbol{X}_{g_1,t}; w_{g_1}) \\ h_{2,t} &= \text{LSTM}(Y_{g_2,t}, \boldsymbol{X}_{g_2,t}; w_{g_2}) \\ &\dots \\ h_{g_J,t} &= \text{LSTM}(Y_{g_J,t}, \boldsymbol{X}_{g_J,t}; w_{g_J}) \end{aligned} \quad (18)$$

where $w_{g_1}, w_{g_2}, \dots, w_{g_J}$ are the trainable weight of LSTM.

Beyond LSTM (Hu and Zheng, 2020), GRU (Dey & Salem, 2017), RNN, and CNN are also used to learn the temporal pattern of the series in our experiments. Through the analysis in the previous section, the difference among RNN, LSTM, and GRU is that the latter two can learn both short-term and long-term dependence of the series and avoid the gradient vanishing. Thus, we obtain the latent representation of the series. In fact, these recurrent layers help us to obtain the nonlinear features of input series with exogenous time series automatically, denoted as $\{h_{g_1,t}, h_{g_2,t} \dots, h_{g_J,t}\}$.

*4.3 Learning representation from spatial patterns*

Considering the fact that various MTSs may play different roles in the process of a specific target series forecasting, the representations of multiple LSTMs may have different weights for the final forecasting. Inspired by the idea of the attention mechanism of human brains regarding how to deal with massive amounts of visual and audio data, we also use the attention layer to learn the weight of each LSTM. Taking a group $g$ as an example, the set of representations learned by multiple multivariate LSTM is $\{h_{g_1,t}, h_{g_2,t}, \dots, h_{g_J,t}\}$. We first use the concatenation operation to combine $\{h_{g_1,t}, h_{g_2,t}, \dots, h_{g_J,t}\}$ to obtain $\boldsymbol{h}_{g,t} = [h_{g_1,t}|h_{g_2,t}|\dots|h_{g_J,t}]$, where "|" is the concatenation operation that splices two matrices together. Then, the attention layer (Vaswani, et al., 2017) is used to learn the attention weight and obtain the global representation. The multiple attention layers are used to learn the weight of multiple representations for a specific series, formulated as

$$\begin{aligned} h_{g_1,t}^a &= \text{attention}(\boldsymbol{h}_{g,t}; \gamma_{g_1}) \\ h_{g_2,t}^a &= \text{attention}(\boldsymbol{h}_{g,t}; \gamma_{g_2}) \\ h_{g_J,t}^a &= \text{attention}(\boldsymbol{h}_{g,t}; \gamma_{g_J}) \end{aligned} \quad (19)$$

where $\boldsymbol{\gamma}_g = \{\gamma_{g_1}, \gamma_{g_2}, \dots, \gamma_{g_J}\}$ are the trainable parameters or the attention weights, and $h_{g_1,t}^a$, $h_{g_2,t}^a, \dots, h_{g_J,t}^a$ are the output of the attention layer. From the last expression of the attention layer in (14)-(16), we can see that in each attention layer, attention(·) is used to learn the different combinations of latent variables $\boldsymbol{h}_{g,t}$ to some extent. If a series contributes more to the final forecasting, it may have a greater weight than others; otherwise, it obtains a little weight.

*4.4 Prediction*

In this step, we take the outputs of the attention layer as inputs to train the prediction model. The objective of series prediction is to reconstruct the relationship between input and output. A one-layer feedforward neural network is used for the prediction function, and it performs as linear regression. For the $j$th MST, let $h_{g_j,t}^a$ be the representations learned by the former attention layer, the whose predicted value is calculated by $\hat{Y}_{g_j,g+h} = \sigma\left(h^a_{g_j,t}W_{g_j,o} + b_{g_j,o}\right)$. Thus the multiple outputs of a set of series are given by

$$\begin{aligned}\hat{Y}_{g_1,t+h} &= \sigma(h^a_{g_1,t}W_{g_1,o} + b_{g_1,o})\\ \hat{Y}_{g_2,t+h} &= \sigma(h^a_{g_2,t}W_{g_2,o} + b_{g_2,o})\\ \hat{Y}_{g_J,t+h} &= \sigma\left(h^a_{g_J,t}W_{g_J,o} + b_{g_J,o}\right)\end{aligned} \quad (20)$$

where $W_{g_j,o}$ and $b_{g_j,o}$ are trainable parameters or the coefficients of linear regression; and $\sigma(\cdot)$ is an identity function; and, $\hat{Y}_{g_1,t+h}$, $\hat{Y}_{g_2,t+h}$ and $\hat{Y}_{g_J,t+h}$ are the predicted value of the series.

*4.5 Loss function of multiple linear regression*

In our study, we want to forecast a set of series simultaneously, which involves multiple responses. Thus, let $\{Y_{g_1,t+h}, Y_{g_2,t+h}, \dots, Y_{g_J,t+h}\}$ be a set of test series, and $\{\hat{Y}_{g_1,t+h}, \hat{Y}_{g_2,t+h}, \dots, \hat{Y}_{g_J,t+h}\}$ be the set of the corresponding predicted series, whose loss function is calculated by the mean square error (MSE) as follows:

$$\text{MSE} = \frac{1}{JH}\sum_{j=1}^{J}\sum_{h=1}^{H}\left(Y_{j,t+h} - \hat{Y}_{j,t+h}\right)^2 \quad (21)$$

where $J$ is the number of MTSs we want to forecast simultaneously and $H$ is the length of predicted values of the target series.

*4.6 Implementation*

We implement the new method to Keras (Ketkar, 2017). To simplify the study, LSTM contains two layers, and the number of neurons in the first and second hidden layers are 16. A one-layer feedforward neural network working as linear regression is used as the prediction model or the output layer, whose number of hidden neurons is the same as the number of outputs. For example, in a group $g$, which contains $m_g$ regions, the number of hidden neurons in the last layer is equal to $m_g$. The Adam algorithm (Kingma & Ba, 2014) with a learning rate of 0.005 is applied to estimate the unknown weight matrix and bias vectors. Considering the size of the real datasets, we set the batch size to 1 in Section 4. In contrast, in Section 5, the training epochs are set to 50 and the batch size is set to 64. Based on these settings, the algorithm of the proposed method is described as follows. Similarly, GRU, RNN and CNN are applied to learn the temporal pattern replacing LSTM in the new method in our analysis.

**Algorithm 1. Temporal-Spatial dependencies ENhanced deep learning model (TSEN) for forecasting household leverage**

---

**Inputs**: Set of multiple MTSs $\{Y_{j,t}, X_{j,t}\}_{j=1}^{J}$.

**Outputs**: Set of forecasted values $\{\hat{Y}_{1,t+h}, \hat{Y}_{2,t+h}, \dots, \hat{Y}_{J,t+h}\}$ with $h$-steps.

**Procedure**:

**1** Input of multiple MTSs $\{Y_{j,t}, X_{j,t}\}_{j=1}^{J}$.

**2** Clustering and screening by the hierarchical clustering algorithm.

**2a** Calculate the similarities of provinces and call the clustering algorithm to divide the provinces into *K* groups, where each group contains multiple MTSs $\{Y_{g_1,t}, X_{g_1,t}\}, \ldots, \{Y_{g_J,t}, X_{g_J,t}\}$.

**3** For each group $\{Y_{g_j,t}, X_{g_j,t}\}$, use multiple MTSs to train the proposed method.

**3a** Feed multiple MTSs into the new deep learning framework.

**3b** Learn the new representations of multiple MTSs in parallel by multiple LSTMs.

**3c** Learn the attention scores and combination of representations of multiple LSTMs.

**3d** Forecast the household leverage of the corresponding multiple regions.

**4** Output $\{\hat{Y}_{g1,t+h}, \hat{Y}_{1,t+h}, \hat{Y}_{2,t+h}, \ldots, \hat{Y}_{J,t+h}\}$ using the proposed method.

## 5. Application

Financial time series forecasting is imperative to the computational intelligence field among finance researchers from both academia and the financial industry due to its broad scope of application and substantial influence. Financial time series forecasting is a valuable tool to foresee whether a particular nation will suffer from financial risk in the next couple of months or days so that the governers can make a good fiscal policy to hedge against market risks and optimize investments in a nation in advance. In this section, we model and learn the time series of household debt in China which are reported and collected monthly.

### 5.1 Data collection

Studies on world economic and financial history in the last three decades show that the rapid growth of credit and the sharp rise of leverage cause systemic risks. Especially, the rise of household debt is often the fuse of an outbreak of financial crisis (Clarke, 2019; Mian, et al., 2017). The degree of household leverage determines an economy's vulnerability to a financial crisis. It turns out that such a crisis can spur strong fluctuations in house prices and consumption of goods (Hintermaier & Koeniger, 2018). Without governmental intervention, excessive household borrowing will lead to a vicious cycle of "debt deflation" with tighter borrowing constraints, reduced consumption, risen unemployment, and fallen asset values. This will eventually lead to a financial crisis and cause a long-term economic recession (Berisha & Meszaros, 2017; Boz & Mendoza, 2014; Dong & Xu, 2020). An important reason for the outbreak of the global financial crisis in 2008 was the excessive leverage in the financial system. And, excessive credit contributed to the outbreak of the debt crisis for highly leveraged households (Aalbers, 2015). For example, U.S. household leverage sharply increased in those years before the current economic recession. The dramatic and absolute rise in U.S. household leverage from 2002 to 2007 is unprecedented, looking back on the past 25 years. By the same token, since the outbreak of the COVID-19 pandemic in February 2020, the major global economy entities have made large-scale loose monetary policies, which has a negative spillover effect of high debt and high inflation. And then, these entities have reached a global consensus on curbing the excessive expansion of household debt. It indicates that the increasing household leverage may reveal an economic recession. If the next financial crisis is once again triggered by events that we did not foresee, it will be disastrous and seriously impede economic recovery during the era of the COVID-19 pandemic.

Despite the fact that China's economy is entering a new stage of development, the debt risk accumulated by the rapid growth of China's household leverage has attracted people's attention: it becomes a potential danger to economic operation and financial security. According to the report of the Bank for International Settlements, "China's household leverage ratio is 60.3% greater than that of the world's at the end of 2020, 2.4% higher than that of the G20 as a whole, and is quite similar to that of Japan and the European Union; at the same time, it is close to the 60% level of household debt in Japan before the economic bubble burst in the 90s of last century." A similar trend can be found in the United States. According to the congressional research service survey, "during 2020, different types of consumer debt—consisting of mortgages, credit cards, auto loans, and student loans—have exhibited different patterns during the COVID-19 pandemic. Notably, credit card balances declined sharply in the second quarter by about $76 billion, the largest quarterly decline on record. Mortgage debt increased, and other household debt remained relatively flat." Although interest payments generate a flow of revenue from indebted households to financial institutions, the consequences of such debt-based financialization system remain under-explored. Local governments need to pay attention to regulating and controlling the scale of household debt and make policies to redistribute revenue and mitigate the wealth inequality.

Indeed, accurate forecasting of household leverage may help local governments to make reasonable and effective decisions or policies, especially when the future development is uncertain. Household leverage is obtained by dividing household debt balance by the gross domestic product (*GDP*). And, household debt is defined as all liabilities of households (including non-profit institutions serving households) that require payments of interest or principal to creditors at fixed dates in the future. The household debt is a sum of numberical values of the following liabilities: loans (primarily mortgage loans and consumer credit) and other accounts payable. The scale of household leverage is highly related to the macroeconomic cycle, monetary policy, real estate market, potential population size, and social consumption. During the economic expansion period, long-term low-interest rates and tax reduction policies will stimulate the expansion of the household leverage (Campbell & Hercowitz, 2009; Canakci, 2021; Clarke, 2019; Del Rio & Young, 2008). Hence, some exogenous time series such as *Population*, *CPI*, *Disposable income per capita*, *M2 scale*, *Average selling price of commercial housing*, and *Total retail sales of social consumer goods* are imperative to the forecast of the trend of household leverage in the future.

*5.2 Data description*

We conducted experiments on household leverage forecasting to test the performance of the proposed model. The dataset consists of 24 MTSs with 72 months: 23 provinces and 1 nation from January 2015 to December 2020. the household leverage is The leverage of the household sector (calculated by household debt divided by GDP). Household debt is a monthly time series from the database of the People's Bank of China; GDP (100 million CNY) is the gross national product. All the values of predictive variables are collected from the National Bureau of Statistics, China and the Wind database, including *Population* (10 thousand CNY) as yearly data, *CPI* (taking the same month last year as 100) as monthly data, *Disposable income per capita* (1 CNY) as quarterly data, *M2 scale increment* (100 million CNY)—the increment of China's broad money—as monthly data, *Average selling price of commercial housing* (1 CNY per square meter) as yearly data, and *Total retail sales of social consumer goods* (cumulative growth) as monthly data. More, *average selling price of commercial housing* and *Total retail sales of social consumer goods* are collected from the National Bureau of Statistics, China, while other predictors are obtained from the Wind database. The trends of household leverage in China from January 2015 to December 2020 are shown in Fig.3.

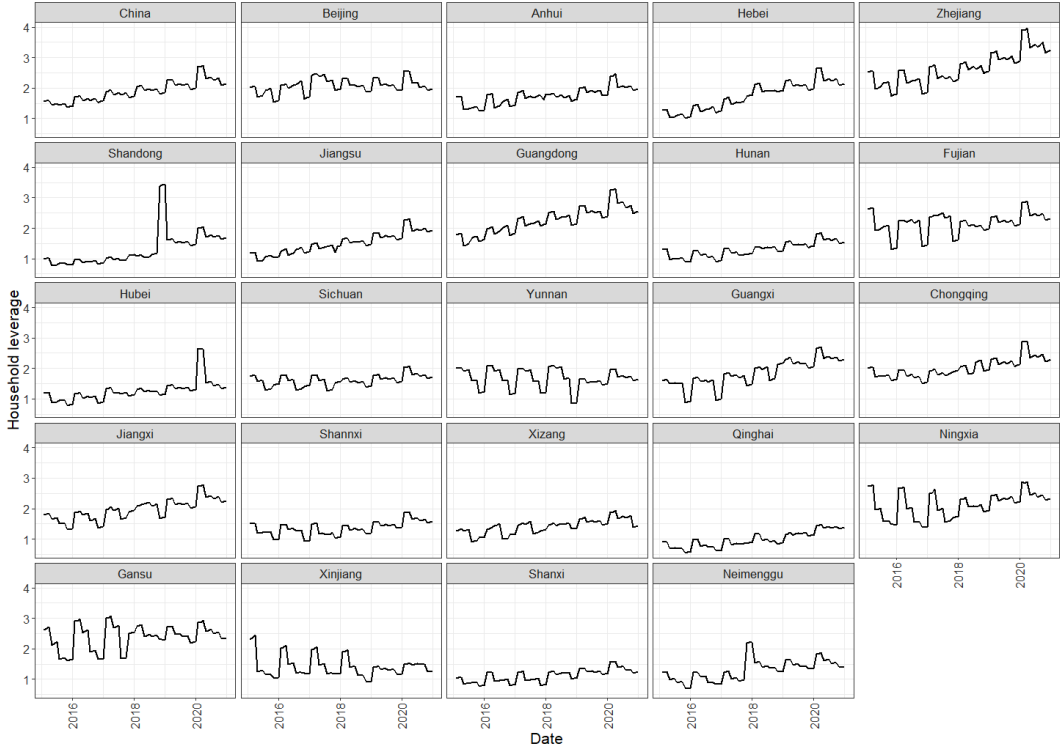

**Fig.3. Trends of household leverage in China from January 2015 to December 2020**

*5.3 Experimental settings*

The experiments conducted in the previous section are based on the analysis of datasets on the *Keras* platform (Ketkar, 2017). Before feeding the input data to the new neural networks, all the multiple MTSs are normalized beforehand. Each MTS consists of several exogenous time series. These variables are used to predict the target time series for the next period or even future period.

We use the out-of-samples performance estimate procedures, where a time series is split into two parts: an initial fit period in which a model is trained, and a subsequent (temporally) testing period held out for estimating the loss of that model (Cerqueira, Torgo, & Mozetič, 2020). For instance, the convolutional neural network (CNN), RNN, LSTM, GRU, and the proposed new method are trained by the training multiple MTSs $\{Y_{j,t}, \boldsymbol{X}_{j,t}\}_{j=1}^{J}$ and then they forecast the future values of multiple targets $\{\hat{Y}_{1,t+h}, \hat{Y}_{2,t+h}, \ldots, \hat{Y}_{J,t+h}\}$. Two metrics are used to measure the performance of the model: mean absolute error (MAE), which is one of the most common metrics used to measure the forecasting accuracy based on 100 repeatedly random experiments, and root MSE (RMSE), which weighs the average squared difference between the estimated value and the actual value. For each selection of training set, we shuffle the dataset randomly and independently. Let $\{Y_{1,t+h}, Y_{2,t+h}, \ldots, Y_{J,t+h}\}$ and $\{\hat{Y}_{1,t+h}, \hat{Y}_{2,t+h}, \ldots, \hat{Y}_{J,t+h}\}$ be the true values and predictive values respectively, the MAE and RMSE are calculated as

$$\text{MAE}_j = \frac{1}{N} \sum |\hat{Y}_{j,t+h} - Y_{j,t+h}| \qquad (22)$$

and

$$\text{MSE}_j = \frac{1}{N} \sum (\hat{Y}_{j,t+h} - Y_{j,t+h})^2 \qquad (23)$$

where $N$ is the length of the time series we want to forecast for $j = 0,1,2,\ldots,J$.

*5.4 Results of forecasting household leverage of 23 provinces*

In this study, we use the new method to forecast the household debt leverage of some provinces in China, which includes 72 observations, 70% of which are included in the training data, and the remaining ones are used as testing data. According to the framework of the new method, the process of analysis contains two steps as follows.

**Step 1. Clustering and screening.**

Before forecasting the household leverage of multiple provinces, we put the provinces into six groups according to the similarity of any two different temporal sequences through a clustering algorithm to avoid overfitting. The method for measuring the similarity between any two time series is called timestep-based similarity, which uses the Euclidean distance to reflect point-wise temporal similarity. For example, Beijing, Anhui, and Hebei are in the same group because of their correlated household debt leverage calculated by the training series. The clustering result is shown in Fig.4, and more details are presented in Table 1.

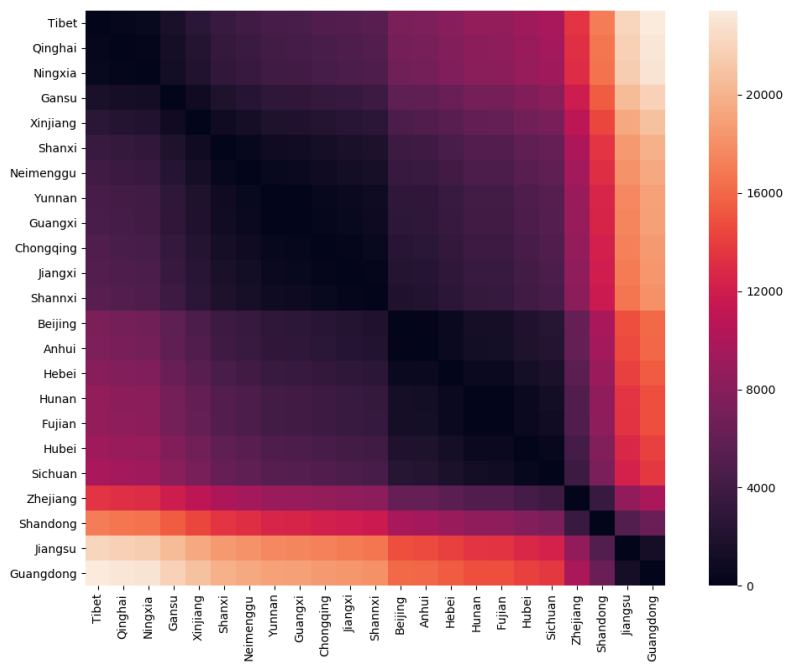

**Fig.4. Heatmap of the similarities of 23 provinces' household leverage**

**Table 1. Groups of provinces**

| Group | Names of provinces |
|---|---|
| 1 | Beijing, Anhui, Hebei |
| 2 | Zhejiang, Shandong, Jiangsu, Guangdong |
| 3 | Hunan, Fujian, Hubei, Sichuan |
| 4 | Yunnan, Guangxi, Chongqing, Jiangxi, Shannxi, |
| 5 | Tibet, Qinghai, Ningxia |
| 6 | Gansu, Xinjiang, Shanxi, Neimenggu |

*Note: Following eight provinces are not included in our analysis because of some missing information - Hainan, Tianjing, Jilin, Heilongjiang, Shanghai, Guizhou, Liaoning, and Henan.*

**Step 2. Modeling and learning multiple time series cluster by cluster.**

We compare the performance of the new method with other alternative methods for forecasting future values of multiple time series cluster by cluster. And then, for each cluster of provinces, we predict the household leverage of multiple provinces in the next third month ($Y_{j,t+3}$) by only looking at these correlated provinces in the same cluster. The former forecasting task is relatively easier than the latter. We forecast the household leverage in the next third month ($Y_{j,t+3}$) for two reasons. One reason is that governments need time to draw up policies, which usually takes one quarter. Another reason is that many statistics are released quarterly in China, and thus the decision-making body of governors always follows after the publication of statistical data. We use both the average predicted RMSE and MAE to show various performances of all methods while forecasting household leverage of 23 provinces in China, reported and listed in Tables 2-3.

Table 2. Average RMSE of forecasting household leverage of multiple provinces in the next third month ($Y_{j,t+3}$) based on the proposed method and alternative methods

| Provinces | TSEN-GRU | TSEN-LSTM | TSEN-RNN | TSEN-CNN | GRU | LSTM | RNN | CNN |
|---|---|---|---|---|---|---|---|---|
| Beijing | 0.0708 | 0.0600 | 0.1116 | 0.0675 | 0.0662 | 0.0613 | 0.0711 | 0.0803 |
| Anhui | 0.0742 | 0.0665 | 0.1225 | 0.0907 | 0.0734 | 0.0746 | 0.0893 | 0.0931 |
| Hebei | 0.0712 | 0.0616 | 0.1346 | 0.0977 | 0.0667 | 0.0663 | 0.0910 | 0.0951 |
| Zhejiang | 0.1180 | 0.1092 | 0.1785 | 0.1300 | 0.1682 | 0.1441 | 0.1633 | 0.1558 |
| Shandong | 0.0667 | 0.0644 | 0.1733 | 0.1477 | 0.0670 | 0.0649 | 0.0934 | 0.0945 |
| Jiangsu | 0.0584 | 0.0559 | 0.0867 | 0.0839 | 0.0633 | 0.0661 | 0.1053 | 0.0905 |
| Guangdong | 0.0831 | 0.0863 | 0.1330 | 0.1382 | 0.0885 | 0.0887 | 0.1148 | 0.1127 |
| Hunan | 0.0713 | 0.0598 | 0.0796 | 0.0736 | 0.0484 | 0.0495 | 0.0643 | 0.0418 |
| Fujian | 0.1200 | 0.1165 | 0.1381 | 0.1689 | 0.1315 | 0.0924 | 0.1043 | 0.1188 |
| Hubei | 0.1599 | 0.1536 | 0.2048 | 0.1768 | 0.1573 | 0.1596 | 0.1901 | 0.1763 |
| Sichuan | 0.0917 | 0.0682 | 0.0748 | 0.1050 | 0.0602 | 0.0599 | 0.0657 | 0.0677 |
| Yunnan | 0.1160 | 0.0753 | 0.1272 | 0.1193 | 0.0923 | 0.0512 | 0.1049 | 0.0751 |
| Guangxi | 0.0603 | 0.0552 | 0.0851 | 0.1550 | 0.0714 | 0.0685 | 0.0922 | 0.0780 |
| Chongqing | 0.0950 | 0.0907 | 0.1061 | 0.1388 | 0.1099 | 0.1139 | 0.0946 | 0.1016 |
| Jiangxi | 0.0781 | 0.0572 | 0.0867 | 0.1017 | 0.0630 | 0.0607 | 0.0732 | 0.0794 |
| Shannxi | 0.0721 | 0.0675 | 0.0724 | 0.0655 | 0.0976 | 0.0933 | 0.0863 | 0.0785 |
| Xizang | 0.0601 | 0.0513 | 0.0751 | 0.0788 | 0.0530 | 0.0528 | 0.0669 | 0.0647 |
| Qinghai | 0.0526 | 0.0506 | 0.0733 | 0.0728 | 0.0463 | 0.0447 | 0.0563 | 0.0531 |
| Ningxia | 0.0681 | 0.0659 | 0.1055 | 0.0931 | 0.0743 | 0.0684 | 0.0830 | 0.0911 |
| Gansu | 0.0838 | 0.0580 | 0.0696 | 0.1129 | 0.0911 | 0.0756 | 0.0981 | 0.0807 |
| Xinjiang | 0.0784 | 0.0612 | 0.1248 | 0.1281 | 0.1037 | 0.0737 | 0.0821 | 0.0872 |
| Shanxi | 0.0287 | 0.0270 | 0.0474 | 0.0578 | 0.0407 | 0.0419 | 0.0555 | 0.0520 |
| Neimenggu | 0.0430 | 0.0396 | 0.1271 | 0.1082 | 0.0477 | 0.0471 | 0.0758 | 0.0648 |

Table 3. Average MAE of forecasting household leverage of multiple provinces in the next third month ($Y_{j,t+3}$) based on the proposed method and alternative methods

| Provinces | TSEN-GRU | TSEN-LSTM | TSEN-RNN | TSEN-CNN | GRU | LSTM | RNN | CNN |
|---|---|---|---|---|---|---|---|---|
| **Beijing** | 0.1742 | 0.1338 | 0.2506 | 0.1635 | 0.1663 | 0.1391 | 0.1725 | 0.1963 |
| **Anhui** | 0.1736 | 0.1506 | 0.2739 | 0.2135 | 0.1583 | 0.1616 | 0.2044 | 0.2180 |
| **Hebei** | 0.1809 | 0.1349 | 0.2905 | 0.2309 | 0.1474 | 0.1363 | 0.2057 | 0.2201 |
| **Zhejiang** | 0.2649 | 0.2324 | 0.3735 | 0.2890 | 0.4149 | 0.3347 | 0.3661 | 0.3501 |
| **Shandong** | 0.1527 | 0.1409 | 0.3560 | 0.3090 | 0.1436 | 0.1415 | 0.2225 | 0.2309 |
| **Jiangsu** | 0.1287 | 0.1228 | 0.1980 | 0.1954 | 0.1252 | 0.1322 | 0.2338 | 0.1957 |
| **Guangdong** | 0.2107 | 0.2198 | 0.2989 | 0.3253 | 0.2301 | 0.2020 | 0.2620 | 0.2509 |
| **Hunan** | 0.1748 | 0.1497 | 0.1802 | 0.1621 | 0.1213 | 0.1230 | 0.1496 | 0.0995 |
| **Fujian** | 0.2979 | 0.2964 | 0.2838 | 0.3575 | 0.3300 | 0.2246 | 0.2531 | 0.2727 |
| **Hubei** | 0.2940 | 0.2797 | 0.4381 | 0.3835 | 0.2864 | 0.2808 | 0.3939 | 0.3499 |
| **Sichuan** | 0.2365 | 0.1758 | 0.1796 | 0.2375 | 0.1419 | 0.1387 | 0.1566 | 0.1519 |
| **Yunnan** | 0.2837 | 0.1962 | 0.2729 | 0.2569 | 0.2229 | 0.1399 | 0.2200 | 0.1833 |
| **Guangxi** | 0.1507 | 0.1466 | 0.1973 | 0.3351 | 0.1855 | 0.1780 | 0.2240 | 0.1862 |
| **Chongqing** | 0.2198 | 0.2175 | 0.2415 | 0.2972 | 0.2533 | 0.2676 | 0.2130 | 0.2256 |
| **Jiangxi** | 0.2081 | 0.1545 | 0.1979 | 0.2329 | 0.1695 | 0.1559 | 0.1900 | 0.1955 |
| **Shannxi** | 0.1881 | 0.1753 | 0.1707 | 0.1543 | 0.2631 | 0.2487 | 0.2147 | 0.1842 |
| **Xizang** | 0.1569 | 0.1353 | 0.1658 | 0.1724 | 0.1391 | 0.1429 | 0.1657 | 0.1563 |
| **Qinghai** | 0.1457 | 0.1420 | 0.1788 | 0.1849 | 0.1281 | 0.1209 | 0.1493 | 0.1469 |
| **Ningxia** | 0.1806 | 0.1642 | 0.2502 | 0.2356 | 0.1906 | 0.1706 | 0.2090 | 0.2185 |
| **Gansu** | 0.2105 | 0.1420 | 0.1694 | 0.2676 | 0.2192 | 0.1857 | 0.2366 | 0.2057 |
| **Xinjiang** | 0.1816 | 0.1496 | 0.2711 | 0.2712 | 0.2727 | 0.1923 | 0.1997 | 0.2006 |
| **Shanxi** | 0.0692 | 0.0713 | 0.1124 | 0.1324 | 0.1038 | 0.1053 | 0.1307 | 0.1230 |
| **Neimenggu** | 0.1014 | 0.0927 | 0.2691 | 0.2392 | 0.1166 | 0.1153 | 0.1786 | 0.1586 |

Tables 2-3 show that the proposed new framework with LSTM for forecasting household leverage of multiple provinces in the next third month ($Y_{j,t+3}$) outperforms all alternative methods in most provinces. Both RMSE and MAE of the new framework with LSTM are almost smallest among RMSE and MAE of all methods. LSTM is also good at forecasting household leverage in the next third month by only using single MTS but exhibit poorer performance than the proposed method. This result illustrates that taking into account the temporal and spatial dynamics of MTSs can enhance the performance of the methods. More, it also indicates that the spatial pattern enlarges the signal for predicting individual household leverage because of introducing correlated MTSs. Especially, the stronger the correlation between the MTSs, the more accurate the proposed framework is. In the next section, we will illustrate this result in the simulation study. This is the reason why we need to screen the highly correlated MTSs to build the forecasting model. Without clustering and screening, the forecasting accuracy will be decreased while introducing uncorrelated and redundant MTSs.

## 5.5 Results of forecasting household leverage in China

Similarly, we also forecast household leverage at the national level. As before, to avoid the proposed model's overfitting, we only select a few provinces to help us to predict the household leverage of China. These regions have a large scale of household debts and significantly contribute to the national household debt. From Table 4, we can see that the first three provinces Guangdong,

Zhejiang, and Jiangsu, have the largest household debt. Therefore, these three provinces are employed to forecast the household leverage of the entire nation. Besides, all the analysis of the forecasting results of the household leverage of China are presented in Table 5.

Table 4. Scale and proportion of household debt in various provinces

| Provinces | Scale (billion) | Percentage | Cumulative percentage | Provinces | Scale (billion) | Percentage | Cumulative percentage |
|---|---|---|---|---|---|---|---|
| Guangdong | 929122.7 | 13.06% | 13.06% | Yunnan | 128404 | 1.81% | 76.13% |
| Zhejiang | 668047.8 | 9.39% | 22.45% | Tianjin | 106462.1 | 1.50% | 77.63% |
| Jiangsu | 617695.5 | 8.68% | 31.14% | Neimenggu | 82588.3 | 1.16% | 78.79% |
| Shandong | 389085.8 | 5.47% | 36.61% | Shanxi | 72208.39 | 1.02% | 79.81% |
| Fujian | 327072.7 | 4.60% | 41.21% | Gansu | 69583.05 | 0.98% | 80.78% |
| Shanghai | 288367 | 4.05% | 45.26% | Jilin | 63915.5 | 0.90% | 81.68% |
| Sichuan | 265535.2 | 3.73% | 48.99% | Xinjiang | 59062.87 | 0.83% | 82.51% |
| Hebei | 250199.9 | 3.52% | 52.51% | Heilongjiang | 52333.6 | 0.74% | 83.25% |
| Anhui | 241921 | 3.40% | 55.91% | Hainan | 35938.83 | 0.51% | 83.75% |
| Beijing | 232738.7 | 3.27% | 59.18% | Ningxia | 29231.53 | 0.41% | 84.16% |
| Hubei | 211279.2 | 2.97% | 62.15% | Qinghai | 12571.51 | 0.18% | 84.34% |
| Hunan | 205575.1 | 2.89% | 65.04% | Xizang | 9473.311 | 0.00% | 84.47% |
| Jiangxi | 185775.2 | 2.61% | 67.66% | Guizhou | | | 84.47% |
| Chongqing | 184096.7 | 2.59% | 70.24% | Henan | | | 84.47% |
| Guangxi | 159565 | 2.24% | 72.49% | Liaoning | | | 84.47% |
| Shannxi | 130845.6 | 1.84% | 74.33% | | | | |

Table 5. Average RMSE and MAE of forecasting household leverage of China in next third months ($Y_{j,t+3}$)

| Values | Measures | TSEN-GRU | TSEN-LSTM | TSEN-RNN | TSEN-CNN | GRU | LSTM | RNN | CNN |
|---|---|---|---|---|---|---|---|---|---|
| $Y_{j,t+3}$ | RMSE | 0.0831 | 0.0757 | 0.1061 | 0.1237 | 0.0876 | 0.0808 | 0.1023 | 0.0965 |
| | MAE | 0.1812 | 0.1602 | 0.2370 | 0.2732 | 0.1877 | 0.1667 | 0.2350 | 0.2098 |

Table 5 indicates that the proposed method outperforms all the alternative methods on both MAE and RMSE. It illustrates that household leverage in multiple regions can enlarge the signal for predicting household leverage at a national level and help to improve the performance of the model. Thus, the experimental results reveal that a single deep learning model (such as LSTM, GRU, RNN, and CNN) can work well only if it just considers the independent information of the predicted province. By contrast, although the proposed method includes the information of other correlated provinces for modeling, the inclusion potentially improves the forecasting performance of household leverage of a target province. At the same time, the clustering algorithm selects and puts useful information into the model, which helps to avoid the influence of noise on prediction results and improve model performance.

*5.6 Statistical comparisons of the applied models*

To access whether the performance of the proposed approach is much better than that of the other methods, it is necessary to carry out a statistical test. To perform the statistical test, we compare all the performances in terms of MAE and RMSE as it is a popular accuracy measure in the context of time series forecasting for predicting household leverage in different provinces. Two statistics are employed to test the results. One is the Friedman test, which is conducted to determine whether there are significant differences between the concerning method and others through observing MAE and RMSE. The null hypothesis of the test is that there are no differences among the compared algorithms. If the test results can reject the hypothesis, then we can conduct the Wilcoxon signed-rank test to figure out whether the prosed model is much better than the best-compared method among all the alternative methods.

The Friedman test compares the average ranks of different methods. For the average rank of method $j$, $AR_j$ is the arithmetic average of its ranking in different data sets. And, Friedman statistic of the Friedman test is computed by Eq (11)

$$\chi_F^2 = \frac{12D}{k(k+1)} \left[ \sum_{j=1}^{k} AR_j^2 - \frac{k(k+1)^2}{4} \right] \qquad (24)$$

where D is the number of datasets, and k represents the number of methods for comparison. And, $\chi_F^2$ is the chi-squared distribution with k-1 degrees of freedom, shown in table 6. Table 6 shows the p-values of all tables that are significant, where there are differences among all the compared models' performances.

Table 6. The performance difference test of all the compared models by the Friedman test

| Tables | Table4 | Table5 |
|---|---|---|
| Statistics | 125.2174 | 111.142 |
| p-value | 0.00 | 0.00 |

In the next step, we use Wilcoxon signed-rank test to figure out whether there are any significant differences among the proposed method and the other methods. Tables 2-3 show that the proposed method with LSTM is better than the others, and LSTM is the relatively best method among the alternatives. Based on these results, we further compared these two methods by the Wilcoxon signed-rank test to show their statistical differences in table 7.

Table 7. The performance difference between TSEN-LSTM and LSTM by
Wilcoxon signed-rank test (the null hyperthesis is TSEN-LSTM is better than LSTM)

| Tables | Table4 | Table5 |
|---|---|---|
| Statistics | 75 | 91 |
| p-value | 0.0281 | 0.0786 |

Table 7 shows that there are significant differences in the ranks between TSEN-LSTM and LSTM. It means that the prediction ability of TSEN-LSTM is better than LSTM in forecasting the values of household leverage in the next third month, but they have a similar performance in forecasting the values in the next month. Obviously, LSTM is relatively good so it always has the suboptimal ranking. Although the proposed model of this paper is the optimal prediction model, the gap between this model and the suboptimal model is stable. This is the reason why the result of post hoc test between these two models is statistically insignificant. Therefore, it is still clear that our model is robust and stable.

## 6. Simulation

Other than the application of the new model, we still need to know under what kind of conditions the model performs well. For instance, the performance varies when it models and learns multiple series with different spatial correlation strengths. To answer the question about how correlation strength influences model fitting, we generate different artificial time series by the following settings with different correlation strengths for the simulation studies. Besides, the artificial time series is also used in our experiments to illustrate the influence of the sample size.

### 6.1 Simulation settings

Let $Y_{k,t} = (y_t, y_{t-1}, \ldots, y_1)$ be a target time series and $X_{k,t} = \{X_{1t}, X_{2t}, \ldots, X_{mt}\}$ be the exogenous MTS with $m$ series. To simplify the simulation studies, we use $Z_{k,t} = \{Y_{k,t}, X_{k,t}\}$ to represent this $m + 1$ dimensional MTS. Suppose the linear model for the conditional mean of the data generation process (DGP) of the observed series may be of the finite order VARMA process, such as

$$Z_{k,t} = \Phi_{k,1} Z_{k,t-1} + \Phi_{k,2} Z_{k,t-2} + \cdots + \Phi_{k,t-p} Z_{k,t-p}$$
$$-\Gamma_{k,t} - \Theta_{k,1} \Gamma_{k,t-1} - \cdots - \Theta_{k,q} \Gamma_{k,t-q} \quad (25)$$

where $\Phi_{k,1}, \Phi_{k,2}, \ldots, \Phi_{k,t-p}$ are $(m+1) \times (m+1)$ autoregressive parameters matrices while $\Theta_{k,1}, \Theta_{k,t-1}, \ldots, \Theta_{k,t-q}$ are moving average parameter matrices also of dimension $(m+1) \times (m+1)$, and $p$ and q are the orders of autoregressive process and moving average process, respectively. In addition, $\Gamma_{k,t} = \{u_{1t}, u_{2t}, \ldots, u_{(m+1)t}\}$ are white-noise with zero mean, nonsingular, time-invariant $E(\Gamma_{k,t} \Gamma'_{k,t}) = \Sigma_\Gamma$, and zero covariance $E(\Gamma_{k,t} \Gamma'_{k,t-h}) = 0$, where $h = \pm 1, \pm 2, \ldots$.

Given the parameters of the VARMA process, both the main part of a target series $Y_{k,t}$ and the spatial correlation of any two target series $Y_{s,t}$ and $Y_{v,t}$ are generated by the VARMA process separately. For instance, we say a $m + 1$ dimensional MTS $Z_{s,t} = \{Y_{s,t}, X_{s,t}\}$ is generated by the VARMA process. Also, $Y_{k,t}$ can be represented by its lag terms and exogenous series $X_{k,t}$. Besides, the spatial parts $Y'_{s,t}$ and $Y'_{v,t}$ of two target series $Y_{s,t}$ and $Y_{v,t}$ are also generated in the same way. Thus, the final series $Y^f_{s,t}$ is the sum of the spatial parts $Y'_{s,t}$ and the main part $Y_{s,t}$.

More specifically, our simulation contains four MTSs and each of them contains one target series and five endogenous series generated by the VARMA(3,3) process. All entries of $\Phi_{k,1}, \Phi_{k,2}, \Phi_{k,3}$ and only the diagonal elements of $\Theta_{k,1}, \Theta_{k,t-1}, \ldots, \Theta_{k,t-q}$ are set to random numbers sampled from uniform distribution $U(0.5, 0.5)$ randomly and independently while the other elements of $\Theta_{k,1}, \Theta_{k,t-1}, \ldots, \Theta_{k,t-q}$ are set to 0. Besides, the non-diagonal entries of $\Sigma_\Gamma$ are set to 0.7 while the non-diagonal entries are set to 2. In addition to generating four MTSs independently, the spatial correlation of four MSTs is also generated by the VARMA(3,3) process with the same settings.

Considering the number of observations and noisy series that may influence the performance of forecasting models, we use three cases to demonstrate the forecasting performance in different numbers of observations and noisy series. The number of observations is set to 100, 1000, and 100 in case 1, case 2 and case 3, respectively; each case contains four MTSs including a target series and 5 exogenous series. Besides case 3 and case 4 contain 5 noisy exogenous series generated by standard normal distribution N(0,1) while others do not have any noisy series. All of these settings are shown in Table 8.

Table 8. Different cases of simulation studies

| Cases | # of obs. | # of MTSs | # of correlated exogenous series | # of uncorrelated series |
|---|---|---|---|---|
| Case 1 | 100 | 4 and correlated | 5 | 0 |
| Case 2 | 1000 | 4 and correlated | 5 | 0 |
| Case 3 | 100 | 4 and correlated | 5 | 5 |
| Case 4 | 100 | 4 and uncorrelated | 5 | 5 |

## 6.2 Results

We compare the performance of the new method with those of other alternative methods, such as CNN, RNN, LSTM, and GRU. All the results are based on simulation datasets generated by following the setting of table 8. RMSE and MAE of forecasting are shown in tables 9-10.

Table 9. Average RMSE of forecasting values of artificial data
in next third months ($Y_{j,t+3}$) based on the proposed method and the alternative methods

| Case | Provinces | TSEN-GRU | TSEN-LSTM | TSEN-RNN | TSEN-CNN | GRU | LSTM | RNN | CNN |
|---|---|---|---|---|---|---|---|---|---|
| Case 1 | Region1 | 0.3800 | 0.3882 | 0.5680 | 0.6425 | 0.3952 | 0.4002 | 0.4292 | 0.4870 |
| | Region2 | 0.3760 | 0.3935 | 0.3957 | 0.4431 | 0.3795 | 0.4130 | 0.5127 | 0.4362 |
| | Region3 | 0.3765 | 0.3931 | 0.4833 | 0.5692 | 0.3834 | 0.4076 | 0.6227 | 0.4701 |
| | Region4 | 0.3791 | 0.3922 | 0.6332 | 0.5147 | 0.3906 | 0.4099 | 0.5544 | 0.6181 |
| Case 2 | Region1 | 0.5015 | 0.8002 | 4.8606 | 7.3620 | 0.5212 | 0.9611 | 1.8912 | 2.2228 |
| | Region2 | 0.4666 | 0.9652 | 6.5017 | 1.5077 | 0.4312 | 1.0908 | 1.9582 | 1.0096 |
| | Region3 | 0.3775 | 0.6628 | 4.2459 | 3.2078 | 0.4652 | 0.8751 | 3.7594 | 2.6217 |
| | Region4 | 0.5451 | 0.6860 | 1.6977 | 6.2885 | 0.4475 | 1.2250 | 2.1011 | 2.6994 |
| Case 3 | Region1 | 0.3750 | 0.3758 | 0.4132 | 0.4203 | 0.3779 | 0.3825 | 0.4073 | 0.3973 |
| | Region2 | 0.3761 | 0.3751 | 0.4289 | 0.4229 | 0.3798 | 0.3837 | 0.4092 | 0.4087 |
| | Region3 | 0.3773 | 0.3765 | 0.4180 | 0.4156 | 0.3767 | 0.3862 | 0.4087 | 0.4163 |
| | Region4 | 0.3778 | 0.3766 | 0.4293 | 0.4228 | 0.3809 | 0.3848 | 0.3964 | 0.4003 |
| Case4 | Region1 | 0.2612 | 0.2646 | 0.2806 | 0.2861 | 0.2673 | 0.2874 | 0.2845 | 0.2790 |
| | Region2 | 0.2616 | 0.2655 | 0.2944 | 0.2888 | 0.2680 | 0.2865 | 0.2878 | 0.2890 |
| | Region3 | 0.2619 | 0.2626 | 0.2856 | 0.2879 | 0.2706 | 0.2932 | 0.2893 | 0.2813 |
| | Region4 | 0.2624 | 0.2617 | 0.2856 | 0.2876 | 0.2659 | 0.2892 | 0.2803 | 0.2836 |

Table 10. Average MAE of forecasting values of artificial data
in the next third month ($Y_{j,t+3}$) based on the proposed method and the alternative methods

| Case | Provinces | TSEN-GRU | TSEN-LSTM | TSEN-RNN | TSEN-CNN | GRU | LSTM | RNN | CNN |
|---|---|---|---|---|---|---|---|---|---|
| Case 1 | Region1 | 2.0854 | 2.1395 | 2.4422 | 2.4483 | 2.1510 | 2.2079 | 2.2561 | 2.3691 |
| | Region2 | 2.0835 | 2.1336 | 2.2130 | 2.2808 | 2.0981 | 2.2392 | 2.3765 | 2.2999 |
| | Region3 | 2.0879 | 2.1354 | 2.3917 | 2.4259 | 2.1058 | 2.2224 | 2.4690 | 2.3637 |
| | Region4 | 2.0915 | 2.1285 | 2.4688 | 2.3656 | 2.1267 | 2.2310 | 2.4029 | 2.4401 |
| **Case 2** | **Region1** | 1.9998 | 2.2856 | 4.8620 | 5.4588 | 2.0376 | 2.4864 | 3.0186 | 3.2168 |
|  | **Region2** | 1.9740 | 2.4128 | 4.4098 | 2.6210 | 1.9994 | 2.5965 | 3.0747 | 2.4646 |
|  | **Region3** | 1.9119 | 2.1726 | 4.3334 | 3.5510 | 2.0012 | 2.4145 | 4.3699 | 3.3589 |
|  | **Region4** | 2.0354 | 2.1791 | 2.7807 | 5.2585 | 2.0062 | 2.6845 | 3.1115 | 3.3384 |
| **Case 3** | **Region1** | 1.9218 | 1.9103 | 2.1269 | 2.1498 | 1.9269 | 1.9502 | 2.0739 | 2.0256 |
|  | **Region2** | 1.9210 | 1.9024 | 2.1996 | 2.1542 | 1.9349 | 1.9521 | 2.0975 | 2.0864 |
|  | **Region3** | 1.9225 | 1.9068 | 2.1075 | 2.1193 | 1.9134 | 1.9848 | 2.1062 | 2.1278 |
|  | **Region4** | 1.9272 | 1.9041 | 2.1702 | 2.1531 | 1.9331 | 1.9600 | 2.0329 | 2.0884 |
| **Case4** | **Region1** | 1.2210 | 1.2404 | 1.3287 | 1.3348 | 1.2539 | 1.3403 | 1.3212 | 1.3090 |
|  | **Region2** | 1.2234 | 1.2452 | 1.3671 | 1.3522 | 1.2565 | 1.3428 | 1.3398 | 1.3582 |
|  | **Region3** | 1.2232 | 1.2290 | 1.3483 | 1.3410 | 1.2676 | 1.3744 | 1.3511 | 1.3111 |
|  | **Region4** | 1.2272 | 1.2337 | 1.3359 | 1.3471 | 1.2446 | 1.3419 | 1.3215 | 1.3320 |

The results in tables 9-10 show that, when forecasting multiple regions in the next third month ($Y_{j,t+3}$), the proposed model has no obvious advantages over the alternative models in the case of small samples (case1). However, when dealing with large samples (case 2), the method proposed in this paper are relatively stable and have the smallest prediction error in most cases, in terms of both MAE and RMSE. Besides, in the case of small samples, if the series is not correlated with each other, or if the exogenous variables are not related, the alternative models that model a single series are better choices over the proposed model. It indicates that when there are many differences between any two series, the joint modeling method shown above in this paper will lead to a certain degree of noise, which affects the accuracy of the prediction of the target. Therefore, we can conclude that our model outperforms other alternatives. Overall, the forecasting results based on the simulation data prove that the proposed method in this paper is suitable for large samples that have a high correlation between any two series, which improves the prediction performance by utilizing the spatial correlation between the related series, thus showing more robustness and effectiveness than the alternatives.

## 7. Discussion and conclusion

This study proposed a new approach for household leverage forecasting using the so-called Temporal-Spatial dependencies ENhanced deep learning model (TSEN). The new method includes a screening and clustering algorithm, multiple deep learning models, an attention layer, and a simple one-layer feedforward neural network for prediction. By the means of clustering, series are divided into different groups according to their relevance. Then, each LSTM is used to learn the representation from temporal patterns of each series, and the attention layer is used to learn the representation from spatial patterns among these series. Finally, the prediction layer can calculate the prediction results through a one-layer feedforward neural network. The new method is used to forecast the household leverage or debt of China and that of several Chinese provinces. The results show that the new approach outperforms other alternative methods. Our experiments reveal that it is a good strategy to predict the time series at time $h$ while considering other related MTSs. It indicates that the other similar time series may be informative for that it amplifies the representation of the predicted time series. The simulation studies also show that correlated series will enhance the performance of forecasting, especially when they are highly correlated. Similar to the mechanism of multiple kernel learning, the new model has good performance on forecasting based on various informative MTSs.

The proposed method can be applied in many fields, such as forecasting global household debt and leverage. Since 2008, the year of the financial crisis, global household debt has been increasing rapidly, which draws the governments' attention around the world. Therefore, the application of the proposed method for this problem can help governments to make reasonable decisions and coordinate the nation's governance for household debt risk prevention. In addition, it could be a useful tool for forecasting global carbon emissions and climate change, which are directly related to macroeconomic variables. Similarly, the household leverage prediction in this paper is based on macroeconomic variables and spatially correlated information. Thus, carbon emissions, climate change, and household leverage prediction may have some common grounds for the applications of the proposed model. Other potential applications of the model include forecasting other macroeconomic indicators, such as the emission trends of various greenhouse gases and the range of global temperature change, and financial indicators, such as the volatility of the S&P 500 (Brandt & Jones, 2006; Huck, 2009), option pricing (Poon & Granger, 2003), commercial decision making in retail (Böse, et al., 2017), and GDP growth rates (Bańbura & Rünstler, 2011; Hoogstrate, Palm, & Pfann, 2000), and other indicators in biological sciences (Stoffer & Ombao, 2012) and medicine (Topol, 2019).